\documentclass[runningheads]{llncs}
\usepackage{graphicx}
\begin{document}
\title{Framework to generate perfusion map from CT and CTA images in patients with acute ischemic stroke: A longitudinal and cross-sectional study}
%
\titlerunning{Framework to generate perfusion map in acute stroke patients}
%
\author{Chayanin Tangwiriysakul\inst{1}\orcidID{0000-0001-5134-0767}  \and
Pedro Borges\inst{1} \and
Stefano Moriconi\inst{1, 3} \and
Paul Wright\inst{1} \and
Yee-Haur Mah\inst{4} \and
James Teo\inst{4} \and
Parashkev Nachev\inst{2} \and
Sebastien Ourselin\inst{1} \and
M. Jorge Cardoso\inst{1}}
\authorrunning{C Tangwiriyasakul, et al.}
%
\institute{School of Biomedical Engineering and Imaging Sciences, King’s College London, London, SE1 7EU, UK \and
UCL Queen Square Institute of Neurology, University College London, London, WC1B 5EH, UK \and 
Support Center for Advanced Neuroimaging (SCAN), University Institute of Diagnostic and Interventional Radiology, University of Bern, Inselspital, Bern University Hospital, 3010 Bern, Switzerland \and  
King’s College Hospital NHS Foundation Trust, Denmark Hill, London, SE5 9RS, UK \\ \email{{chayanin.tangwiriyasakul}@kcl.ac.uk}}
%
\maketitle              
\begin{abstract}
Stroke is a leading cause of disability and death. Effective treatment decisions require early and informative vascular imaging. 4D perfusion imaging is ideal but rarely available within the first hour after stroke, whereas plain CT and CTA usually are. Hence, we propose a framework to extract a predicted perfusion map (PPM) derived from CT and CTA images. In all eighteen patients, we found significantly high spatial similarity (with average Spearman's correlation = 0.7893) between our predicted perfusion map (PPM) and the T-max map derived from 4D-CTP. Voxelwise correlations between the PPM and National Institutes of Health Stroke Scale (NIHSS) subscores for L/R hand motor, gaze, and language on a large cohort of 2,110 subjects reliably mapped symptoms to expected infarct locations. Therefore our PPM could serve as an alternative for 4D perfusion imaging, if the latter is unavailable, to investigate blood perfusion in the first hours after hospital admission.
\end{abstract}
\section{Introduction}
Stroke is one of the leading causes of disability or death. About 15 million people suffer from stroke worldwide per year ~\cite{ref_url028}. The first hour after admission to a stroke unit is considered the golden hour, in which the patients who receive suitable treatments have a higher chance of avoiding long-term brain damage ~\cite{ref_article1_002}. The benefit of receiving proper treatment within this golden hour in stroke was seen in all age groups ~\cite{ref_article1_0022}. Thus, all efforts should be made to give all patients the most suitable treatment within that time frame. Currently, 4-dimensional CT or MR perfusion imaging (4D-CTP) is used to investigate blood flow through the brain vessels and ultimately predict infarct location ~\cite{ref_article1_003}. However, its acquisition process is complex. On top of that, 4D-CTP requires specialised software to compute clinically useful 3D maps of cerebral blood flow, cerebral blood volume, or time-to-maximum (T-max). Taken together, they limit the number of patients eligible for 4D-CTP. In contrast, plain CT and CT angiography (CTA) are routinely acquired on admission to the stroke pathway and so are available earlier and for more patients.

As a neurovascular disease, the loss of brain function after blood vessel blockages cause stroke onset. Three cerebral regions can be defined after stroke onset: the ischemic core, the penumbra and the oligemic region. The ischemic core is the brain area with cerebral blood flow between 4.8 to 8.4 mL/100 g per minute ~\cite{ref_article1_0131,ref_article1_0132}. Unlike the penumbra and the oligemic region, depletion of blood flow in the ischemic core lead to cell death and cause permanent brain damage. Understanding cerebral blood flow using a perfusion map is one of the keys to helping provide patients with proper treatments or predict possible brain damage. 

This study presents a framework to generate a predicted perfusion map (PPM) derived from CT and CTA in the first hours after admission as an alternative to 4D-CTP. We validated our PPM using both longitudinal and cross-sectional analyses. The former was done by comparing the spatial similarity between our PPM and the T-max map derived from 4D-CTP in eighteen subjects. The latter was done by testing the relationship between the PPM and National Institutes of Health Stroke Scale (NIHSS) subscores, which is a standardised series of bedside tests used to assess the severity of stroke symptoms, in a large cohort of 2,110 patients. In summary, we have developed a deployable framework and validated it with clinical data to show it produces images containing similar information about cerebral perfusion to 4D-CTP.
\section{Method}

\subsection{Data set}
We selected a continuous cohort of patients admitted to the stroke unit at King's College Hospital on the basis of having both CT and CTA and evaluation with the NIHSS. Note that right after being admitted to the A\&E unit, NIHSS are evaluated in every patient. Later within 5 to 20 minutes, a CT and CTA are acquired. Inclusion criteria were met by 2,110 patients (mean [SD] age = 68.8 [15.9] years, female = 936 [44\%]). Of these, eighteen had 4D-CTP (mean [SD] age = 61.9 [13.3] years, female = 8 [44\%]). 

\subsection{Data pre-processing and estimation of predicted perfusion map}
Each patient's images were prepared using SPM12 ~\cite{ref_url024}. We coregistered the CT to the CTA, computed affine registration of the CT to MNI space, and applied this to both, reslicing images to 1x1x1 mm resolution. The prepared images were inputs in VTrails ~\cite{ref_article1_004,ref_article1_014}. In this study, we only run the first two steps of VTrails: (1) digital subtraction image pre-processing and (2) vascular contrast enhancement and seeds detection. In step 1, we created a digitally subtracted image (DSA) by composite registration (Affine + BSpline) with CTA as a reference image and CT as the moving image, followed by subtraction of CT from CTA. Later the DSA image was normalised by its maximum value. In step 2, we extracted seed points from the vascular contrast-enhanced version of the DSA ~\cite{ref_article1_004} and ~\cite{ref_article1_014}. In this step, we first applied a gradient anisotropic filter to the DSA image as in Perona-Malik ~\cite{ref_url025,ref_article1_016}. The aim of this filter is to suppress noise while preserving edges (of the vascular structure in our case). We call this filtered DSA image or VSP. We then binarised the VSP image to segment the vessels. Any voxel in the VSP image with an intensity higher than 0.2 is considered a part of the vessel. Note that 0.2 is the default parameter in VTrails ~\cite{ref_article1_004,ref_url026}. The binarised VSP was later converted into the skeleton image (SKEL) using itkBinaryThinningImagheFilter3D in ITK  ~\cite{ref_url027}. The skeleton depicts the centreline of the vascular structure. itkBinaryThinningImagheFilter3D was developed based on Lee et al. ~\cite{ref_article1_017}, which is a 3D decision tree-based algorithm aiming to thin the binary image. Any voxel along the skeleton image with its corresponding value in the VSP image at the same (x,y,z) location greater than its 75th percentile (VTrails' default parameter) was considered a seed point.  

\begin{figure}[h!]
\includegraphics[width=\textwidth]{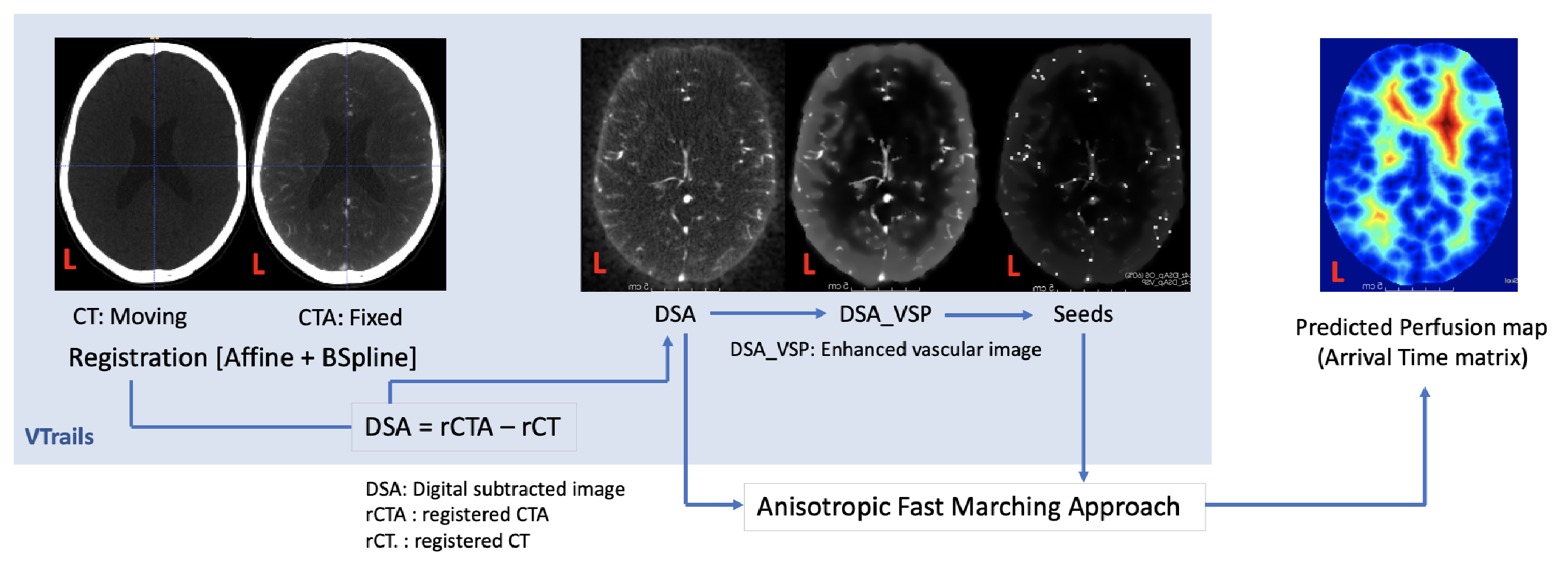}
\caption{Preprocessing steps in VTrails and the estimation of the predicted perfusion map.}\label{Figure1}
\end{figure}

The DSA and Seed images were later used to estimate time-of-arrival at each voxel. A fast marching algorithm was used to estimate the time-of-arrival at each voxel with the seed points as the source and the DSA as the speed potential matrix ~\cite{ref_article1_006}. 
The fast-marching algorithm is a special case of Dijkstra’s algorithm ~\cite{ref_article1_018}. The aim of a fast-marching algorithm is to extract a minimal geodesic path by minimising an energy function weighted by an image speed potential connecting any possible path between two points ~\cite{ref_article1_004,ref_article1_018}. In our study, all the seed points are located along the centerline of the lumen contour. The velocity profile is highest along the centreline and monotonically decreases away from the centre ~\cite{ref_book30}. Thus our seed points located along the centreline are chosen to act as the source to supply blood/ oxygen to its closest neighbouring brain areas (in our study, we assume the continuation of blood flow within the vessels). Each seed point will propagate through the anisotropic medium (in our case, the speed potential matrix). The voxelwise image of time-of-arrival comprises the PPM. Since the DSA was normalised between 0 to 1 (where the voxel intensity inside the vascular structure was higher than its surrounding tissue), the PPM is unitless. A higher voxel value reflects a longer time taken for blood to perfuse the location, hence a higher risk of ischemia at that voxel. Figure ~\ref{Figure1} depicts all steps to estimate the PPM from CT and CTA. We run our scripts using MATLAB2020a on an Intel(R) Core(TM) i9-9900K computer. For each subject, it took 539 seconds (SD=49 seconds) to proprocess CT/CTA images, run VTrails and estimate PPM.


\subsection{Longitudinal analysis}
In eighteen subjects, a 4D-CTP image was acquired. The RAPID-AI software (https://www.rapidai.com/, California USA) was used to derive a Time-to-maximum (T-max) image from the 4D-CTP image. T-max is a parameter of the modelled perfusion representing the time for contrast to reach each voxel from the proximal large artery. It is commonly used to predict infarction maps~\cite{ref_article1_007,ref_article1_008}. In our study, each exported T-max volume was first converted into a greyscale volume and later coregistered to the PPM using SPM-12. To enhance the visibility of the infarct core, we applied a Gaussian filter with a kernel size of 10 voxels to both the exported T-max and the PPM as suggested by Campbell et al. ~\cite{ref_article1_009}. Finally, we estimated a Spearman's rank correlation coefficient between the exported T-max map with the PPM to assess the level of similarity. In this study, we used Spearman's rank correlation coefficient because the unit of the Rapid T-max map and our PPM did not match but are correlated. Furthermore, we evaluated if the spatial similarity between our PPM and T-max correlated with age. In this study, we chose the T-max map as a benchmark since it is used to predict the final infarct size and functional outcomes in patients with ischemic stroke ~\cite{ref_article1_013}. 

\subsection{Cross-sectional analysis}
The NIHSS is a standardised series of bedside tests used to assess the severity of stroke symptoms. It comprises eleven sub-scores such as level of consciousness and limb mobility ~\cite{ref_url1_010}. In this study, we focused on the four primary sub-scores: (1) motor arm left, (2) motor arm right, (3) best gaze, and (4) best language. These were chosen because motor, gaze and language functions have well-established neuroanatomical regions. We tested a general linear model (GLM) with SPM12 using each NIHSS sub-score as a covariate of interest, with age and gender as confounding variables. T-test was used to test for significance with two criteria (1) the p-value $<$ 0.05 (FWE corrected) and (2) the extended threshold $>=$ 100 voxels.  

\section{Results}
\subsection{Longitidinal analysis}
\begin{figure}[h!]
\includegraphics[width=1.0\textwidth]{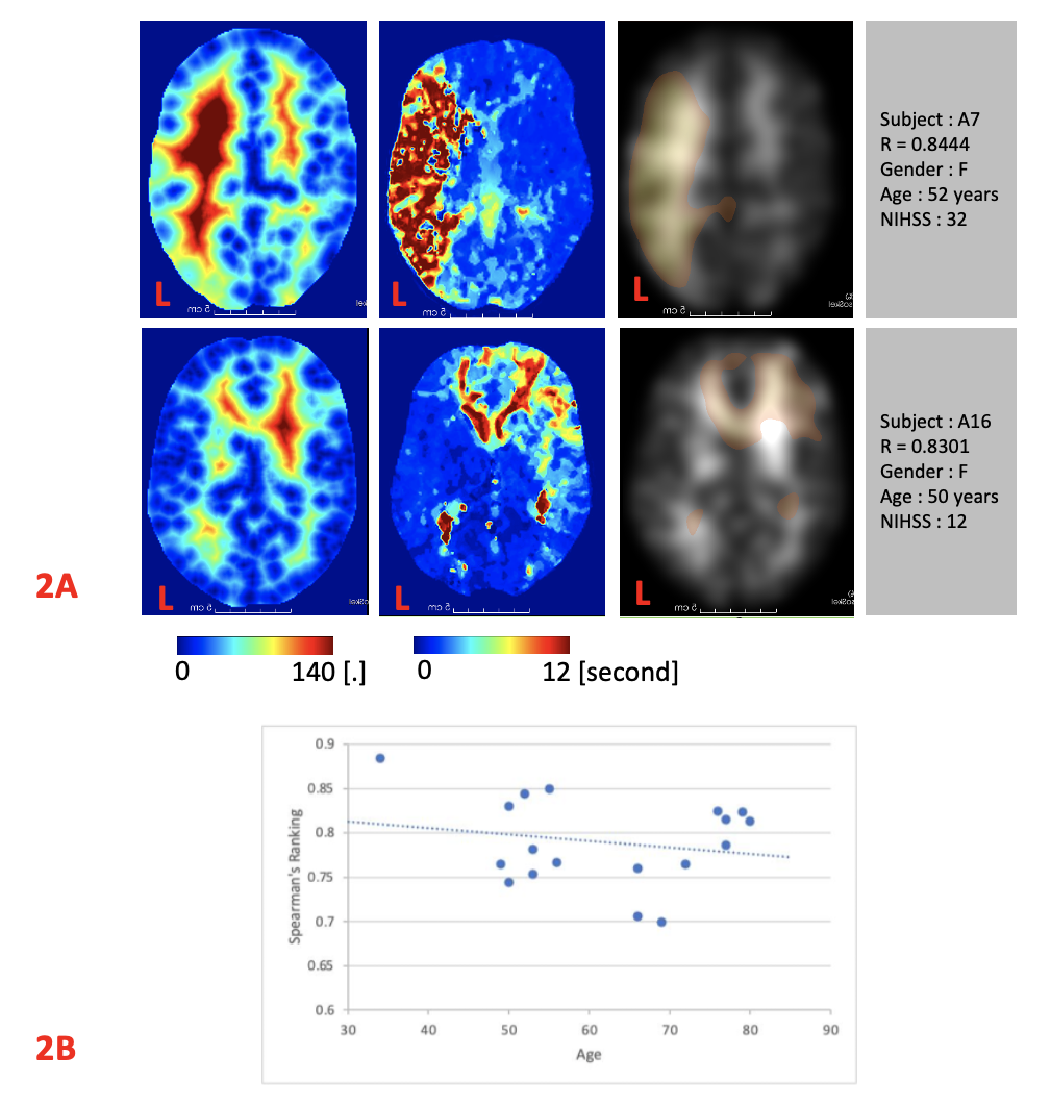}
\centering
\caption{2A: Two typical examples of the PPM and the T-max map. Each row consists of three sub-figures, from left to right, the PPM, the T-max map, and the smoothed T-max map overlay on our PPM's smoothed version. In the predicted perfusion and T-max maps, dark red represents the area with a high risk of permanent brain damage, whereas blue represents the area with a low risk. The overlap map was created for illustrative purposes to highlight the overlap between two brain images. The units in the predicted perfusion and the T-max maps are dimensionless and seconds, respectively. Note that: L = left hemisphere. 2B: Spearman ranks plotted against age for all subjects. Each dot represents each subject's spatial similarity index (Spearman's correlation coefficient). The dotted line shows no significant correlation between age and the spatial similarity index between the two modalities.}
\label{Figure2}
\end{figure}

From the 18 patients, we found an average correlation of 0.7893 (SD=0.049, min=0.6990, max=0.8844). A significant correlation was found in every subject, see Table 1. Figure ~\ref{Figure2}A shows two typical examples. In subject A07, the T-max map and the PPM highlight the ischemic core and penumbra covering almost all of the left hemisphere. In subject A16, the ischemic core located in the right frontal lobe can be seen in both maps. Note that we also estimated the correlation coefficients between the non-smoothed images. Similarly to the smoothed version, both modalities are highly correlated with an average correlation coefficient of 0.707 (SD=0.031, min = 0.659, max = 0.7868). Using Spearman's rank test, we found no significant correlation between the spatial similarity index of PPM-and-Tmax with age (rho = -0.0548, p-value = 0.8290, see Figure ~\ref{Figure2}B, which is desirable as our PPM should be independent of age. 


\subsection{Cross-sectional Analysis}
\begin{figure}[h!]
\includegraphics[width=0.78\textwidth]{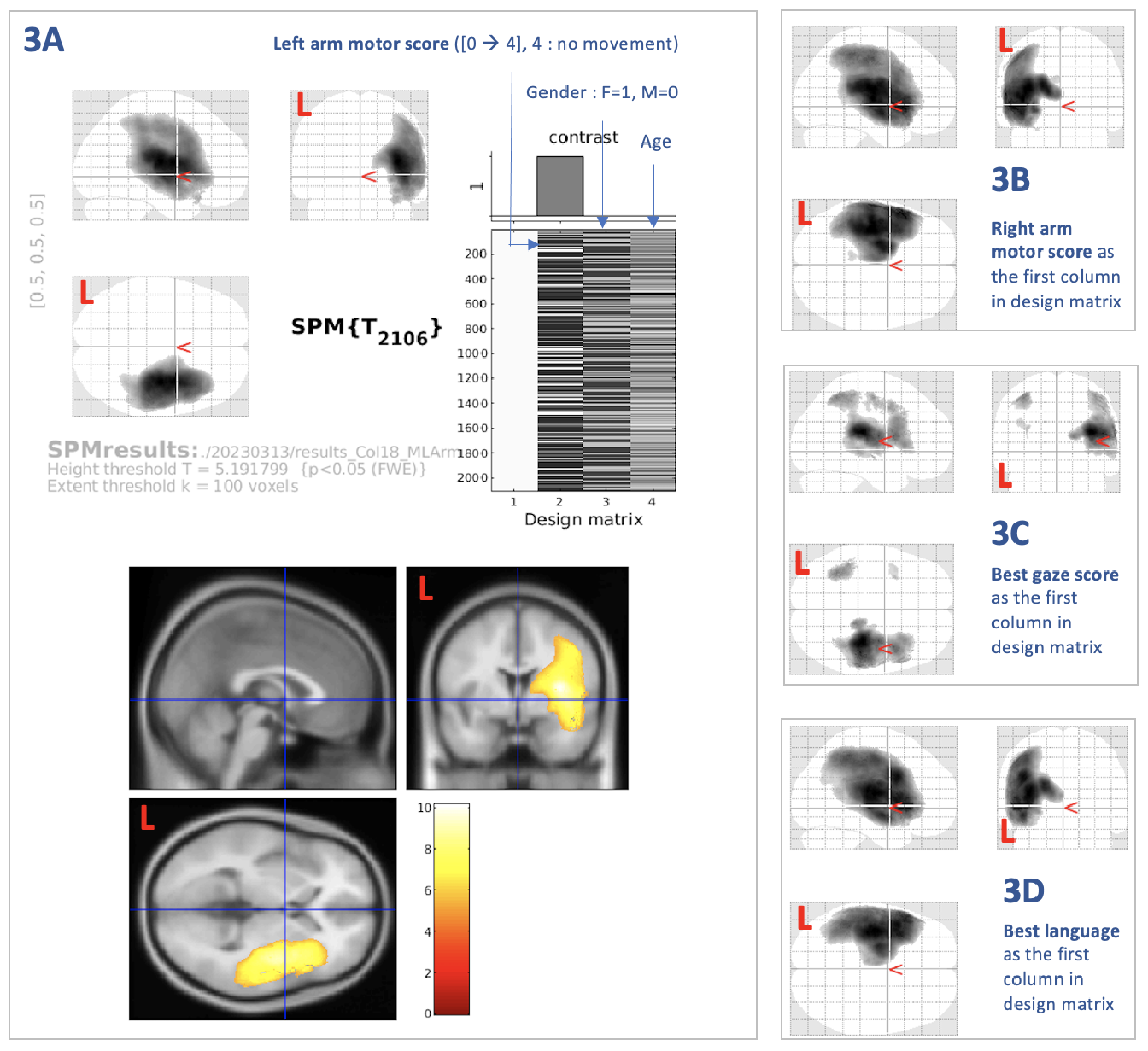}
\centering
\caption{Four GLM models in which different NIHSS subscores were used as covariates. 3A: Left-hand motor score, 3B: Right-hand motor score, 3C: Best gaze score, 3D: Best language score. Each section shows the brain in sagittal, coronal and axial views with a mean intensity projection of the voxelwise significance scores for each GLM contrast (darker = more significant). Section 3A also shows the SPM design matrix with the tested contrast over the variable of interest. L = left hemisphere.}\label{Figure3}
\end{figure}

From the four GLMs derived from 2110 subjects, our PPM mapped brain lesions to stroke symptoms in the expected regions. The left-hand motor score correlated with the right motor cortex and corticospinal tract (Figure ~\ref{Figure3}A; darker regions indicate a more significant effect). As expected, the laterality was reversed for the right-hand motor score (Figure ~\ref{Figure3}B). The best gaze score correlated more to the voxels in the right hemisphere (Figure ~\ref{Figure3}C). The best language score correlated with left perisylvian regions (see Figure ~\ref{Figure3}D). This suggests patients with left hemisphere stroke have a higher chance of having right-hand motor problems as well as problems with languages, moreover, patients with right hemispheric stroke will have a higher chance of having left-hand motor problems and visual problems, both of which match expectations.

\begin{table}
\centering
\caption{Table 1: List of Spearmen's correlation between PPM and Tmax map in the 18 subjects, * = significant correlation, p$<$0.05}\label{dice_exp1}
\begin{tabular}{|c|c|c|c|}
\hline
{\bfseries Subject}  &  {\bfseries Spearmen's correlation} & {\bfseries Subject}  &  {\bfseries Spearmen's correlation}\\
\hline
A1 & 0.7856* & A10 & 0.6990* \\ 
A2 & 0.7529* & A11 & 0.7665* \\
A3 & 0.7817* & A12 & 0.8492* \\
A4 & 0.8237* & A13 & 0.7598* \\
A5 & 0.8146* & A14 & 0.8844* \\
A6 & 0.8131* & A15 & 0.7642* \\
A7 & 0.8444* & A16 & 0.8301* \\
A8 & 0.8238* & A17 & 0.7054* \\
A9 & 0.7648* & A18 & 0.7442* \\
\hline
\end{tabular}
\end{table}

\clearpage

%
%
%
\section{Discussion}
In this work, we presented a framework to generate a perfusion map constrained by physical and geometrical properties as an alternative to the traditional 4D perfusion map. Our PPM is generated from CT and CTA images, which can be acquired at the onset of stroke admission and are commonly available. In the eighteen patients with available 4D-CTP, we confirmed that our PPM was predictive of T-max maps from subsequently acquired 4D-CTP independent of subject age. We chose to compare our PPM with T-max since it is used to define the final infarct size and functional outcomes after recovery ~\cite{ref_article1_013}. Moreover, in the cohort of 2,110 patients, the PPM correlated with stroke symptom scores in predictable, specific regions. Besides the lateralisation found in the contra-lateral motor cortex to the affected hand, we showed symptoms lesions maps associated with language and gaze (visual) problems, which is in line with the findings reported by Bonkhoff et al. ~\cite{ref_article1_015}, which used data collected within one week after stroke onset. 

Like all lesion-symptom mapping analyses, we note that our results are constrained by the underlying vascular anatomy and potential clinical utility in situations where 4D-CTP is impractical. In our patient population, the cohort with plain CT and CTA was over one hundred times larger than the cohort with 4D-CTP. The aim of the work isn't to replace 4D-CTP; it is to provide a means of obtaining as high-fidelity an alternative from admission that relies only on admission scans (CT/ CTA) as opposed to depending on additional scanning sessions. We mathematically derive PPM directly from CT and CTA images, which are the standard scans applied to every patient admitted to a stroke unit. Our PPM, an algorithmically derived estimation of the Tmax map, can provide clinicians with a continuous scale of brain areas with a high risk of cell death with no additional cost of having another brain scan or additional exposure to unnecessary radiation.


\subsection{Conclusion and Future work}
To our knowledge, this is the first time the perfusion map was investigated on this large scale. Our PPM agrees with the T-max map generated from RAPID-AI software and highlights the brain infarction, which emphasizes the predictability power of our PPM. Although the number of patients with RAPID-AI images is still relatively low compared to the number we used in our GLM analysis, a statistically significant correlation between PPM and T-max was found in all subjects. In the future, we will address this limitation by performing the analysis in a large cohort or by using DWI as a surrogate for 4D-CTP, as it is more commonly acquired in hospitals.

\section{Acknowledgements}
CT, PB, SM, PW, PN, SO, and MJC are supported
by the Wellcome Trust (WT213038/Z/18/Z). MJC, JT, and SO are also supported
by the Wellcome/EPSRC Centre for Medical Engineering (WT203148/Z/16/Z),
and the InnovateUK-funded London AI Centre for Value-based Healthcare. YM is supported by an MRC
Clinical Academic Research Partnership grant (MR/T005351/1). JT is also supported by NHSX Ai Award and the Maudsley BRC. PN is also supported by the UCLH NIHR Biomedical Research Centre. 

%
%
%

\end{document}